**Title:** Does Haptic Connection to a Healthy Partner Improve Motor Learning in Chronic Stroke? A Pilot Study of Robot-Based Dyadic Training

**Authors:**

- Erica L. Waters, MSE[1,2]
- Natalia Stroutinsky-Sobotka, MS, OTR/L[3]
- Reem M. Azar, MSE[1,4]
- Frances S. Shofer, PhD[5]
- Pamela Z. Cacchione, PhD, RN, CRNP, BC, FGSA, FAAN[6]
- Rochelle J. Mendonca, PhD, OTR/L, FAOTA, FNAP[7]
- Michelle J. Johnson, PhD[1,2,3,8]

**Corresponding Author:**

- Name: Erica L. Waters
- Address: Rehab Robotics Lab, 1800 Lombard Street, Philadelphia, PA 19146
- Email: waterica@engineering.upenn.edu
- Phone: (215) 893-2695

[1] Rehabilitation Robotics Lab, General Robotics, Automation, Sensing and Perception (GRASP) Laboratory, University of Pennsylvania, Philadelphia, PA, USA
[2] Department of Mechanical Engineering and Applied Mechanics, University of Pennsylvania, Philadelphia, PA, USA
[3] Penn Medicine Rehabilitation, Good Shepherd Penn Partners, Philadelphia, PA, USA
[4] Department of Bioengineering, University of Pennsylvania, Philadelphia, PA, USA
[5] Department of Emergency Medicine, University of Pennsylvania School of Medicine, Philadelphia, PA, USA
[6] School of Nursing, University of Pennsylvania, Philadelphia, PA, USA
[7] Department of Rehabilitation and Regenerative Medicine, Columbia University, New York, NY, USA
[8] Department of Physical Medicine and Rehabilitation, University of Pennsylvania, Philadelphia, PA, USA

**Abstract**

*Purpose:* Robot-based haptic dyads, where two users exchange forces through robots, have been proposed for robot-assisted neurorehabilitation. This intervention could increase motor learning by promoting motor variability, providing an implicit communication channel, and facilitating social interaction. Despite the potential benefits, few studies have investigated robot-based haptic dyads for neurological populations, and none have included a comprehensive clinical assessment.

*Design/Methodology/Approach:* We present a pilot study investigation of individuals with chronic stroke learning a complex tracking task in solo and dyad conditions. Prior to training, participants completed a battery of cognitive, motor, and sensory assessments. We assessed changes in key kinematic outcomes including tracking error, distance traveled, mean absolute power, and sample entropy. We also performed a correlation analysis of clinical scores with baseline tracking errors.

*Findings:* Baseline tracking error was strongly correlated ($r > 0.6$) with lower cognitive scores and higher motor scores. Only the solo group demonstrated lower tracking error at assessment ($\Delta = -5.33$, 95%CI [-10.25, -0.40], $p = 0.027$), but this result was tempered by the removal of an outlier participant. Neither group demonstrated clear retention. The dyad group showed significantly elevated performance, effort, and movement complexity during training.

*Research Limitations/Implications:* The selected task was too difficult to reveal learning for either group, which may be related to its high cognitive demand. Sample size and stroke heterogeneity were also limitation.

*Originality/Value:* Dyadic training increased movement complexity and effort, which are associated with better motor learning. Clinical characteristics such as lesion location, cognition, sensation, and chronicity may influence potential benefits of dyadic training.

## Introduction

Stroke is a leading cause of adult disability in the US. Many people with residual impairments after stroke do not receive sufficient doses of post-stroke neurorehabilitation services to drive neuroplasticity ("Global burden of 369 diseases and injuries in 204 countries and territories, 1990–2019", 2020; Young *et al.*, 2023). Robot-assisted neurorehabilitation can increase dose by enabling more repetitions and decrease the cost of care by enabling clinicians to oversee multiple individuals at once (Gassert and Dietz, 2018; Gower *et al.*, 2025). Beyond motor outcomes, robot-assisted neurorehabilitation has shown promise for promoting cognition, quality of life, and psychosocial functioning (Zanatta *et al.*, 2026). Social isolation and depression are common following stroke (Gu *et al.*, 2025), and addressing these psychosocial issues is essential for sustaining engagement in rehabilitation (Meadmore *et al.*, 2019). Social- and peer-support interventions along with social gaming interventions show promise for improving post-stroke physical and psychosocial outcomes(Lin *et al.*, 2019; Pereira *et al.*, 2021; Thielbar *et al.*, 2020; Wan *et al.*, 2021). Furthermore, multiplayer robot-assisted rehabilitation games are more enjoyable and engaging than solo-games (Novak *et al.*, 2014). Despite the potential for socialization presented by multiuser robotic systems, this aspect has been underexplored.

Robot-based haptic dyads, where two or more users interact through haptically connected robotic devices, have been proposed as a framework for robot-assisted neurorehabilitation that facilitates interactions, leverages social engagement, and provides variable and adaptive assistance (Vianello *et al.*, 2025; Waters and Johnson, 2024). Robot-based haptic dyads have shown promise for motor learning outcomes.

While short-term studies have yielded mixed results (Beckers *et al.*, 2020; Ganesh *et al.*, 2014; Waters and Johnson, 2024), research suggests that long-term training with a haptic connection to a partner can positively affect motor learning and retention outcomes in healthy young adults (Ivanova *et al.*, 2022; Mireles *et al.*, 2017). Among older adults, training with a haptic connection to a partner increased motor learning retention at 1-week post-training (Waters *et al.*, 2026). In the context of neurorehabilitation, pilot studies support that post-stroke participants do not slack in a dyad with a healthy novice and that a haptic connection to human partner may promote motor variability compared to a haptic connection to the target (Short *et al.*, 2025; Waters *et al.*, 2024). In addition to socialization, robot-based haptic dyads may be more effective than target guidance because it introduces a variable human partner (Ivanova *et al.*, 2022; Sabu *et al.*, 2020), and motor variability is essential for motor learning (Herzfeld and Shadmehr, 2014; Wu *et al.*, 2014). Furthermore, a human partner can communicate task goals through haptic interaction (Takagi *et al.*, 2017; Thorne *et al.*, 2019). Robot-based haptic dyads may be particularly useful for individuals with both post-stroke cognitive impairment as well as motor impairment—a group who typically demonstrates worse motor performance and recovery(Bui *et al.*, 2022; Vanbellingen *et al.*, 2025)—by providing implicit instruction via sensorimotor communication with a partner (Hodges and Lohse, 2020; Pezzulo *et al.*, 2019).

Despite these promising results, it remains uncertain whether motor learning is retained across sessions in populations with neurological impairment, as required for neurorehabilitation. Likewise, it is unclear which types of participants would benefit the most from a haptic connection to a healthy partner. While individuals with cognitive impairment could benefit from implicit communication with a partner, previous investigations have only reported motor function, not cognition (Short *et al.*, 2025). Furthermore, participants with sensorimotor deficits might not benefit from the connection to a partner if they are unable to interpret the haptic signals. It is therefore necessary to evaluate motor learning retention within this paradigm while considering an individual's more comprehensive clinical profile.

We present a prospective pilot study investigating motor learning and 1-week retention outcomes of post-stroke participants under two conditions: no haptic feedback and haptic feedback from a healthy partner. We hypothesized that training with a healthy partner would lead to improved performance and learning. We analyze learning and retention outcomes in the context of post-stroke impairments and comment on the influence of a haptic connection to a partner for participants with different levels of post-stroke impairments.

## Methods

### *Participants*

Adults with chronic stroke, defined as at least six months post-stroke, aged 50 and older, who self-reported upper extremity impairment, were recruited to complete the 3-day study. Healthy adults aged 50 and older with no self-reported motor or cognitive impairments were also recruited. All participants completed a clinical evaluation with a licensed occupational therapist (OT). Participants were excluded if they were unable to give informed consent due to greater than moderate cognitive impairment, defined as a score of 12 or less on the Montreal Cognitive Assessment (MoCA) (Nasreddine *et al.*, 2005). Participants were also excluded if they were physically incapable of completing the task, as determined by an OT who assessed if they were able to complete the required arm movement for the task. Finally, participants were excluded in they had greater than mild depression, defined as a score of 4 or more on the Beck's Depression Inventory-Fast Screen (Beck *et al.*, 2000).

***Experimental Procedures***

Participants completed a 3-session protocol, approved by the Institutional Review Board of the University of Pennsylvania (protocol no. 851418; approved on October 10, 2021; located in Philadelphia, PA, USA) according to guidelines set by the Department of Health and Human Services. Figure 1 shows the overall procedure. In the first session, participants completed written informed consent and a clinical assessment battery. In the second session, they learned a robot-based pursuit tracking task. This session consisted of a baseline block (4 trials), a training block (30 trials), and an assessment block (4 trials). Post-stroke participants were assigned to one of two training conditions: (1) solo with no haptic feedback or (2) dyad with a haptic connection to a healthy partner. Healthy participants served as partners to the post-stroke participants in the dyad condition. In the third and final session, participants returned after six to fourteen days and completed a retention block (8 trials) and qualitative interview. Healthy and post-stroke participants completed the same set of procedures with two exceptions: medical verification and Fugl-Meyer Upper Extremity (FM-UE) assessment were not collected for the healthy participants.

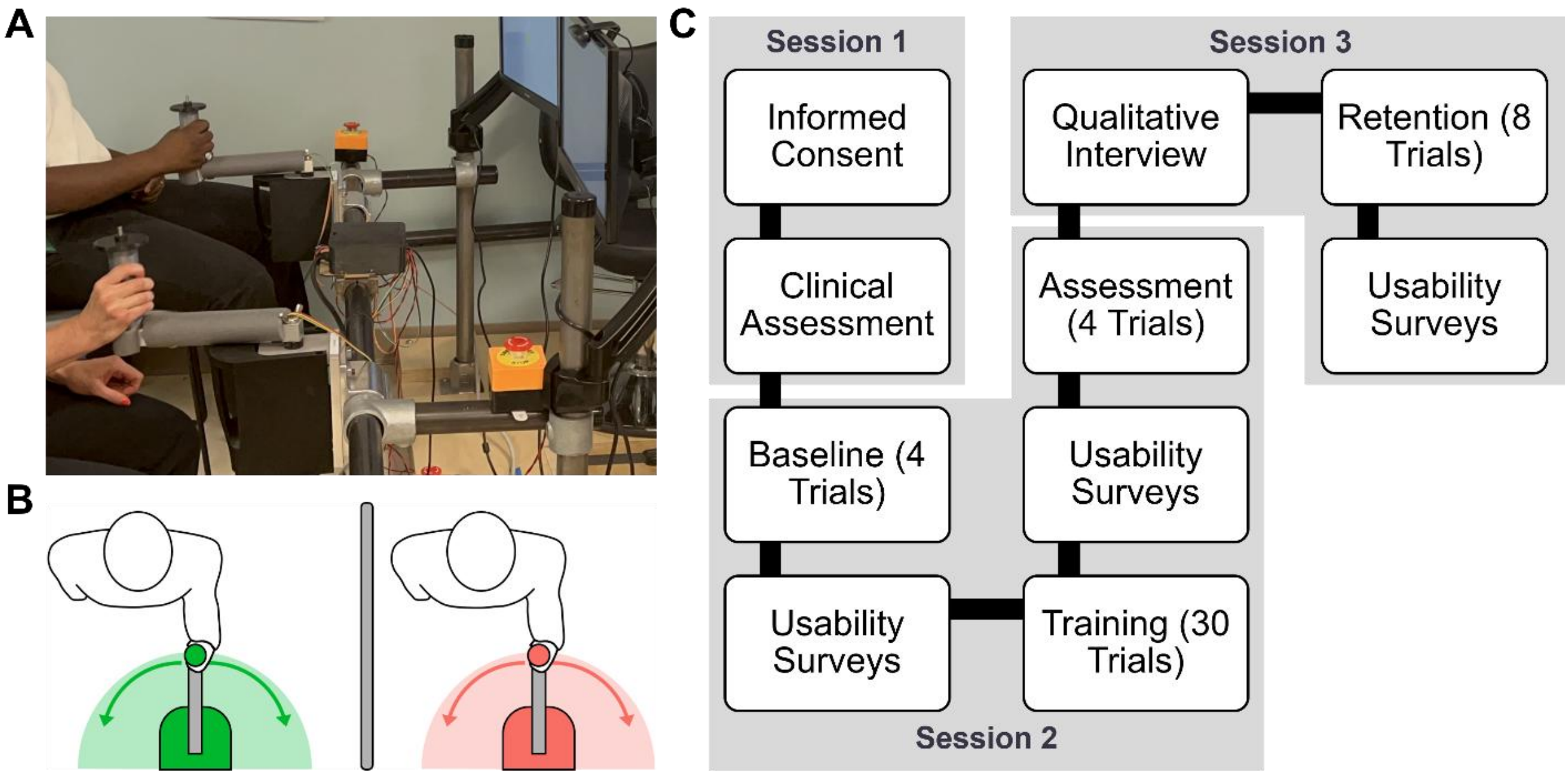

***Figure 1:*** *Experimental methods.* ***A:*** *Set-up of the robot.* ***B:*** *Top-down view of the experimental set-up in the dyadic condition.* ***C:*** *3-day experimental procedure.*

*Medical Verification*

Medical verification was completed via a retrospective review of each post-stroke participant's electronic records through the Epic medical chart information system (Epic Systems, Wisconsin, USA). Stroke site and time were recorded based on the available information from documented stroke events. For each stroke event, the corresponding brain scan was identified, and the radiology reports were used to verify stroke sites. Participants with more than one stroke event were marked as having multiple strokes. In situations where medical verification was not available through Epic, self-report information from the participant was used instead.

*Clinical Assessment*

A licensed OT administered a battery of cognitive, motor, and sensory assessments. The cognitive assessments included Color Trails 1 and 2 (CT1, CT2) and the Montreal Cognitive Assessment (MoCA). CT1 assesses visual attention and sequencing while CT2 assesses selective attention and mental flexibility; performance is defined as completion time (D'Elia *et al.*, 1996). MoCA is a comprehensive cognitive screening scored out of 30 points where a score of 26 or more is considered cognitively intact (Nasreddine *et al.*, 2005). MoCA measures overall cognition but can be separated into subscores for specific cognitive domains. We report the MoCA executive function subscore (MoCA-EF), which is calculated from relevant questions (backward digit span, trail-making test, word similarities, and F-word list generation) and scored out of five points (Lam *et al.*, 2013).

The motor assessments included the Fugl-Meyer Upper Extremity (FM-UE), Box and Block Test (BBT), and the Grooved Pegboard (GP). FM-UE is an assessment of post-stroke motor impairment and is scored out of 66 points with a higher score corresponding to greater function (Fugl-Meyer AR *et al.*, 1975). BBT is a test of motor activity with score determined as the number of blocks transferred in one minute. GP is a fine motor assessment with score determined as the completion time to place 25 grooved pegs into keyed holes (“Grooved pegboard test user instructions”, 2002).

The sensorimotor assessments included the Semmes-Weinstein Monofilament (SWM) and the thumb localizing test (TLT). SWM is an assessment of light touch sensation in which touch sensation is scored according to five grades from “Normal” (5) to “Deep Pressure Sensation Only” (1) where each grade corresponds to the size of filament stimuli that can be perceived with eyes closed (Suda *et al.*, 2021; Weinstein, 1993). SWM grade was defined as the worst recorded grade across all locations (back palm, proximal/distal pinky, distal ring, distal middle, proximal/distal forefinger, distal thumb) for a given hand. TLT is a proprioception test in which the participant moves their opposite hand to their thumb on the assessed side with eyes closed. TLT is graded from 0 to 3 where a score of zero indicates normal proprioception (Suda *et al.*, 2021).

*Robotic Training*

Participants trained on a pursuit tracking task using TheraDyad, a multiuser system consisting of two one-degree-of-freedom robots (Figure 1-left). The robotic interface was configured vertically with the workspace along the user’s transverse plane (Meng *et al.*, 2025). During the dyadic training condition, a virtual spring with a stiffness of 150N/m was rendered to connect the positions of the handles and allow participants to exchange forces; participants were seated beside each other with a partition between their workstations. Participants did not receive any haptic feedback during the baseline, assessment, retention, and solo training condition blocks. The pursuit tracking task involved mapping horizontal, rotational motion of the robot to vertical, linear movement of the cursor and target. For additional feedback, the cursor changed color when inside of the target. Each trial lasted 15 seconds. Participants trained with the more-affected hand. Participants were instructed to follow the target but were not informed of the mapping or source of haptic feedback prior to training.

***Data Analysis***

Outcome Measures

We investigated motor performance across four outcome measures to understand accuracy, efficiency, effort, and movement complexity. Root mean square error (RMSE), defines the error between a participant's location and the target; a larger RMSE value is associated with worse performance (Bui *et al.*, 2022). Normalized distance (ND), representing movement efficiency, is defined as the ratio between total distance traveled by the participant and the total distance required by the task. A ND value below one indicates insufficient distance (Bui *et al.*, 2022). Mean absolute power (MAP), quantifies effort as a function of torque and velocity; higher MAP values are associated with greater effort (Waters *et al.*, 2026). Lastly, sample entropy (SampEn) of position is a measure of complexity and unpredictability (Richman and Moorman, 2000). Higher SampEn values indicate more complex, less predictable movements, and increased SampEn during upper limb training is associated with greater motor learning (Brink *et al.*, 2025). All outcome measures were calculated in Python. SampEn was calculated using the AntroPy library ("antropy", 2026).

Primary Analysis: Linear Mixed-Effects Models

Outcome measures were calculated for all trials and split into means for six time points: baseline (trials 1-4), training 1 (trials 5-14), training 2 (trials 15-24), training 3 (trials 25-34), assessment (trials 35-38), and retention (trials 39-46). We then fit linear mixed-effects models to each outcome measure. Model fixed-effects included group (dyad, solo) and trial (baseline, training 1, training 2, training 3, assessment, retention) as well as the interactions; repeated measures were accounted for with a random-effect of participant. Once models were fit, we conducted type III analysis of variance (ANOVA) to determine the main effects of group, trial, and the interaction between group and trial. As a post hoc analysis, we investigated Tukey-adjusted pairwise differences in estimated marginal means for between- and within-group differences across trials.

*Secondary Analysis: Pearson Correlations*

As a secondary analysis, we calculated Pearson correlation coefficients between the baseline values for all four outcomes and a subset of clinical assessments. Specifically, we selected MoCA-EF, CT1, CT2, FM-UE, BBT, and SWM. The goal of this analysis was to elucidate which skills contributed most to performance on our selected task. Cognitive domains of executive function (MoCA-EF) and attention (CT1) as well as motor function (FM-UE) and activity (BBT) have been shown to relate to tracking performance (Azar *et al.*, 2026; Bui *et al.*, 2022). We further chose to include CT2 given the additional cognitive challenge of visuomotor mapping in our task. We likewise included SWM to see if sensory function was related to performance on the robot-based tracking task. Due to sample size limitations, we calculated Pearson correlation coefficients ($r$) for the combined dyad and solo groups in the baseline block only.

All statistical analyses were performed using R Statistical Software (v4.4.1;R Core Team 2024) (R Core Team, 2024).

## Results

The final study included 10 post-stroke participants (age range of 52-88 yrs, average age 71 ± 11 yrs; 7F/3M), solo (n = 5) and dyad (n = 5), and 5 healthy participants (age range of 54-72 yrs, average age 59 ± 8 yrs; 3F/2M), all dyad (n = 5). Two participants were lost to follow-up before the retention day. Data from these participants was still included, excluding the retention day. Another participant was excluded from the final study due to inadequate motor function for the given task (FM-UE = 11).

### *Demographics and Clinical Profiles*

Demographics and the cognitive and sensorimotor scores between the two groups were similar though the solo group had more individuals with left side hemiparesis than the dyad group (Table I). Only one participant had a measurable proprioception deficit as measured with TLT. For motor scores, FM-UE and BBT were higher for the solo group compared with the dyad group. Healthy participant data is reported in the Supplemental Material.

The type of strokes were heterogenous across the participants with Middle Cerebral Artery (MCA) as the most frequent type of stroke. Three participants had multiple strokes, and they were all in the Solo group; four participants had bilateral lesions, and one of whom was in the dyad group. Time since stroke ranged from less than one year to fourteen years, reflecting variability in chronicity.

**Table I:** Participant Demographics and Stroke Characteristics

| ID | Group | Sex (F/M) | Age (yrs) | Hand (L/R) | MoCA (0-30) | MoCA-EF (0-5) | CT1 (z-score) | CT2 (z-score) | FM-UE (0-66) | BBT (blocks) | TLT (3-0) | SWM (1-5) | Side (L/R/B) | Multiple (Y/N) | Site* | Time Since (yrs) |
|---|---|---|---|---|---|---|---|---|---|---|---|---|---|---|---|---|
| 22 | Solo | M | 63 | R | 24 | 5 | 0.24 | -0.65 | 27 | 8 | 0 | 4 | L | N | PnA | < 1 |
| 24 | Solo | F | 52 | R | 24 | 5 | -3.48 | -0.73 | 26 | 22 | 0 | 4 | B | Y | PnA | 1.5 |
| 31 | Solo | F | 68 | R | 26 | 4 | -0.36 | 0.13 | 32 | 30 | 0 | 4 | L | N | PnA | 14 |
| 39 | Solo | F | 79 | R | 24 | 5 | -2.17 | -3.37 | 56 | 23 | 0 | 1 | B | Y | MCA, PCA, BG | 1 |
| 46 | Solo | F | 76 | L | 17 | 2 | -0.80 | -8.26 | 65 | 44 | 0 | 4 | B | Y | MCA, CERB | < 1 |
| ***Solo Group:*** | | *4F/1M* | *68±11* | *4R/1L* | *23±3* | *4±1* | *-1.31±1.50* | *-2.58±3.44* | *41±18* | *25±13* | *0±0* | *3±1* | *2L/0R/3B* | *3Y/2N* | *2 MCA, 1 PCA, 1 BG, 3 PnA, 1 CERB* | *4±6* |
| 40 | Dyad | M | 59 | L | 27 | 5 | 1.03 | 1.08 | 19 | 0 | 2 | 1 | R | N | MCA, LACI | 1.5 |
| 41 | Dyad | F | 88 | R | 20 | 3 | -0.83 | -1.57 | 58 | 38 | 0 | 3 | L | N | MCA, BG | 8 |
| 47 | Dyad | F | 75 | L | 25 | 4 | -4.12 | -2.47 | 50 | 34 | 0 | 4 | B | N | PnA, CERB | 1.5 |
| 55 | Dyad | F | 74 | L | 23 | 4 | -4.05 | -7.31 | 24 | 0 | 0 | 4 | R | N | MCA, BG | <1 |
| 56 | Dyad | M | 76 | L | 30 | 5 | 0.52 | -0.91 | 21 | 0 | 0 | 1 | R | N | MCA | 5 |
| ***Dyad Group:*** | | *3F/2M* | *74±10* | *1R/4L* | *25±4* | *4±1* | *-1.49±2.46* | *-2.24±3.12* | *34±18* | *14±20* | *0±1* | *3±2* | *1L/3R/1B* | *0Y/5N* | *4 MCA, 0 PCA, 2 BG, 1 PnA, 1 CERB, 1 LACI* | *3±3* |

**Lesion site abbreviations: Middle Cerebral Artery (MCA), Posterior Cerebral Artery (PCA), Basal Ganglier (BG), Pontine Arteries (PnA), Cerebellum (CERB), Lacunar Cerebral Infarct (LACI)*

***Motor Performance Between Group and Across Trials***

Figure 2 shows two examples of post-stroke participants across trials. Qualitatively, we observed increased movement for the dyad participant during training that appears to be partially maintained at assessment and retention. Additional example trajectories are provided in the Supplemental Material. Conversely, for the solo participant, we observed little change in movement following the baseline trials. This pattern was likewise apparent in the overall results. The dyad demonstrated improvement in all four outcomes while connected to a healthy partner during the training trials. The solo group, who did not have haptic assistance, stayed within the baseline 95% confidence interval (Figures 3, Table 2).

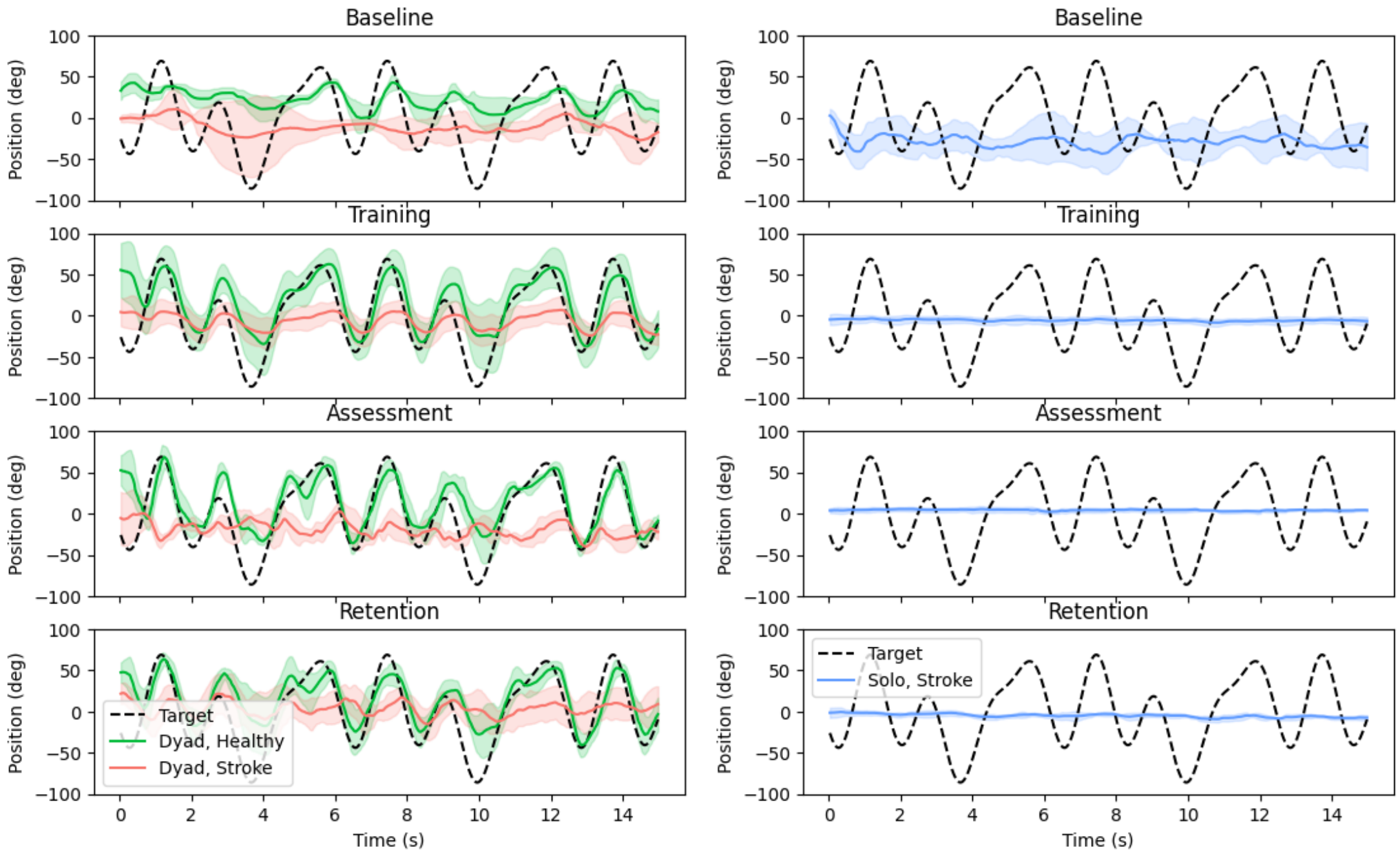

***Figure 2:*** *Example trajectories across trials for two participants with mild motor impairments and moderate cognitive impairments. The dashed lines represent the target trajectory position over time. The solid lines represent mean participant trajectory, and the shaded regions represent standard deviation of trajectories across trials.* ***Left:*** *Dyad post-stroke participant (red, participant 41) paired with a healthy partner (green, participant 43).* ***Right:*** *Solo post-stroke participant (blue, participant 46).*

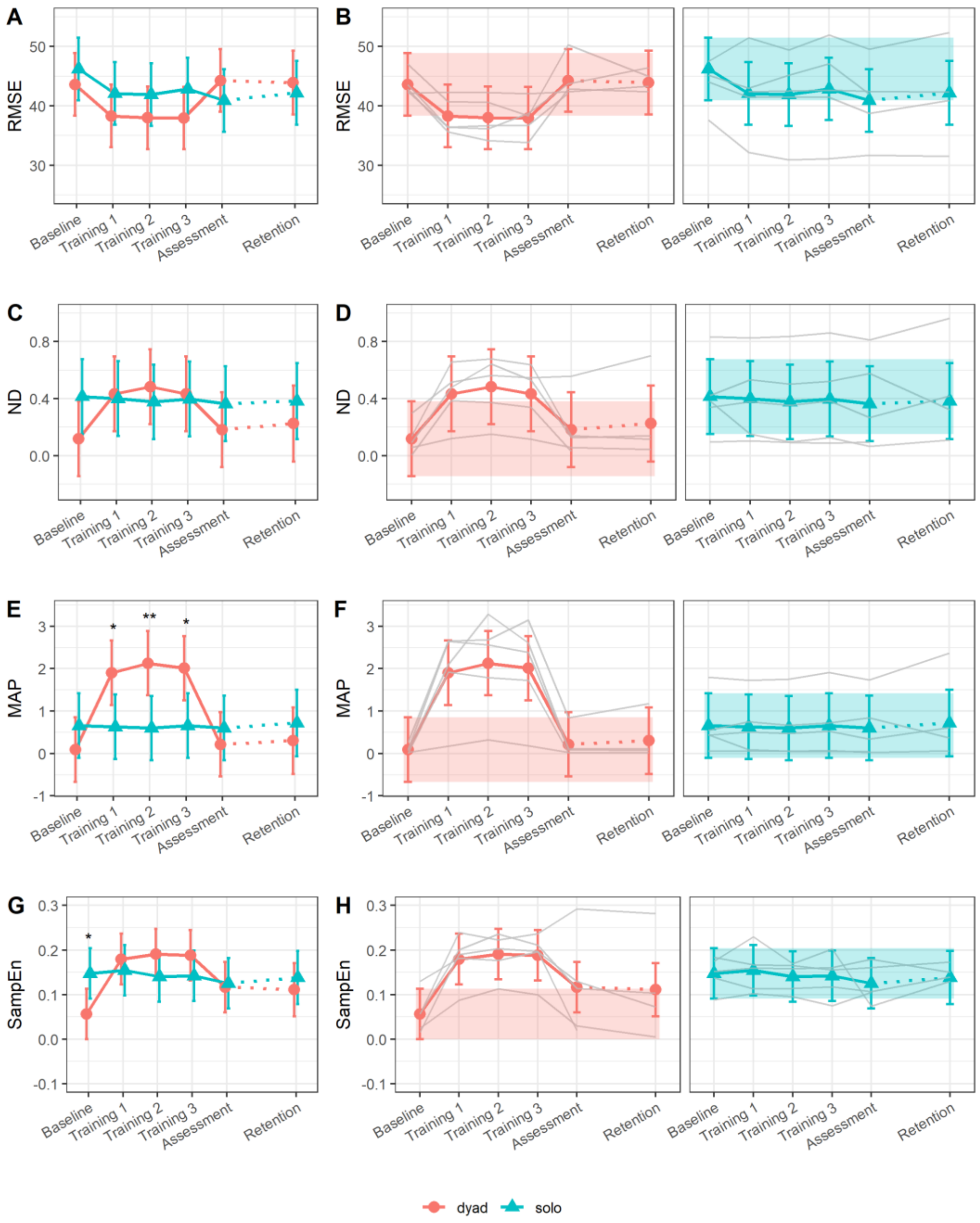

***Figure 3:*** *Estimated marginal means for dyad and solo groups with 95% confidence intervals across six time points. Significant differences between groups are marked (p<0.05: *; p<0.01: **). Shaded areas represent the 95% confidence interval at baseline. Light gray lines represent individual participant values.* ***A-B*** *RMSE comparison between groups and differences across trials compared with baseline. Lower RMSE values indicate more accurate tracking.* ***C-D:*** *ND comparison between groups and differences across trials compared with baseline. Higher ND values indicate greater movement* ***E-F:*** *MAP comparison between groups and differences across trials compared with baseline. Higher MAP values indicate greater effort.* ***G-H:*** *SampEn comparison between groups and differences across trials compared with baseline. Lower SampEn values indicate less movement complexity.*

Table II: Performance Metrics Across Training Day Trials with Change from Baseline

| | | Dyad Stroke | | | Solo Stroke | | |
|---|---|---|---|---|---|---|---|
| **Metric** | **Trial** | **EMM* [95% CI]** | **Δ [95% CI]** | **p-value** | **EMM [95% CI]** | **Δ [95% CI]** | **p-value** |
| *RMSE* | | | | | | | |
| | *Baseline* | *43.6 [38.3, 48.9]* | ---- | | *46.2 [41.0, 51.5]* | ---- | |
| | Training 1 | **38.3 [33.0, 43.6]** | **-5.30 [-10.23, -0.37]** | **0.029** | 42.1 [36.8, 47.4] | -4.14 [-9.06, 0.79] | 0.144 |
| | Training 2 | **38.0 [32.7, 43.3]** | **-5.63 [-10.55, -0.70]** | **0.017** | 41.9 [36.6, 47.2] | -4.32 [-9.25, 0.60] | 0.114 |
| | Training 3 | **37.9 [32.7, 43.2]** | **-5.67 [-10.60, -0.74]** | **0.016** | 42.9 [37.6, 48.1] | -3.37 [-8.30, 1.55] | 0.332 |
| | Assessment | 44.3 [39.0, 49.5] | 0.66 [-4.26, 5.60] | 0.999 | **40.9 [35.6, 46.7]** | **-5.33 [-10.25, -0.40]** | **0.027** |
| | Retention | 43.9 [38.5, 49.3] | 0.30 [-4.98, 5.58] | >0.999 | 42.2 [36.8, 47.6] | -4.04 [-9.32, 1.24] | 0.220 |
| *ND* | | | | | | | |
| | *Baseline* | *0.12 [-0.14, 0.38]* | ---- | | *0.41 [0.15, 0.68]* | ---- | |
| | Training 1 | **0.43 [0.17, 0.69]** | **0.31 [0.10, 0.53]** | **0.001** | 0.40 [0.14, 0.66] | -0.02 [-0.23, 0.20] | 0.999 |
| | Training 2 | **0.48 [0.22, 0.74]** | **0.36 [0.15, 0.58]** | **<0.001** | 0.38 [0.11, 0.64] | -0.04 [-0.25, 0.18] | 0.995 |
| | Training 3 | **0.43 [0.17, 0.70]** | **0.32 [0.10, 0.53]** | **0.001** | 0.40 [0.13, 0.66] | -0.02 [-0.23, 0.20] | >0.999 |
| | Assessment | 0.18 [-0.08, 0.44] | 0.06 [-0.15, 0.28] | 0.948 | 0.36 [0.10, 0.63] | -0.05 [-0.27, 0.17] | 0.980 |
| | Retention | 0.22 [-0.04, 0.49] | 0.11 [-0.12, 0.34] | 0.735 | 0.38 [0.12, 0.65] | -0.03 [-0.26, 0.20] | 0.998 |
| *MAP* | | | | | | | |
| | *Baseline* | *0.09 [-0.67, 0.85]* | ---- | | *0.65 [-0.10, 1.41]* | ---- | |
| | Training 1 | **1.90 [1.14, 2.66]** | **1.81 [0.98, 2.63]** | **<0.001** | 0.63 [-0.13, 1.39] | -0.03 [-0.85, 0.80] | >0.999 |
| | Training 2 | **2.13 [1.37, 2.89]** | **2.04 [1.21, 2.86]** | **<0.001** | 0.60 [-0.16, 1.36] | -0.06 [-0.88, 0.77] | 0.999 |
| | Training 3 | **2.01 [1.25, 2.77]** | **1.92 [1.09, 2.74]** | **<0.001** | 0.66 [-0.10, 1.42] | -0.00 [-0.82, 0.82] | >0.999 |
| | Assessment | 0.21 [-0.54, 0.97] | 0.12 [-0.70, 0.94] | 0.998 | 0.60 [-0.16, 1.36] | -0.06 [-0.88, 0.77] | >0.999 |
| | Retention | 0.30 [-0.48, 1.08] | 0.21 [-0.67, 1.09] | 0.980 | 0.72 [-0.06, 1.50] | 0.06 [-0.82, 0.95] | 0.999 |
| *SampEn* | | | | | | | |
| | *Baseline* | *0.06 [-0.00, 0.12]* | ---- | | *0.15 [0.09, 0.20]* | ---- | |
| | Training 1 | **0.18 [0.12, 0.24]** | **0.12 [0.05, 0.20]** | **<0.001** | 0.15 [0.10, 0.21] | 0.01 [-0.07, 0.09] | 0.999 |
| | Training 2 | **0.19 [0.13, 0.25]** | **0.13 [0.06, 0.21]** | **<0.001** | 0.14 [0.08, 0.20] | -0.01 [-0.09, 0.07] | 0.999 |
| | Training 3 | **0.19 [0.13, 0.24]** | **0.13 [0.05, 0.21]** | **0.039** | 0.14 [0.09, 0.20] | -0.01 [-0.08, 0.07] | >0.999 |
| | Assessment | 0.12 [0.06, 0.17] | 0.06 [-0.02, 0.13] | 0.213 | 0.13 [0.07, 0.18] | -0.02 [-0.10, 0.06] | 0.958 |
| | Retention | 0.11 [0.05, 0.17] | 0.05 [-0.03, 0.14] | 0.388 | 0.14 [0.08, 0.20] | -0.01 [-0.09, 0.07] | 0.999 |

*Estimated marginal means*

*Root Mean Square Error (RMSE)*

The dyad and solo groups showed no differences in RMSE (lower values indicate more accurate tracking) at any time points (Figure 3A) but showed differences within groups at specific trials (Figure 3B). Specifically, the dyad group decreased their RMSE by 5.67 points during the training trials relative to baseline but returned to baseline performance at assessment and retention (Table II). In contrast, the solo group only demonstrated a significant decrease in RMSE at assessment ($\Delta$ =5.33) but returned to baseline performance at retention (Table II).

*Normalized Distance (ND)*

Like RMSE, the dyad and solo groups showed no differences in ND (higher values indicate greater movement) at any time points (Figure 3C) but demonstrated differences within groups at specific trials (Figure 3D). Specifically, the dyad group increased their ND during the training trials relative to baseline but returned to baseline performance at assessment and retention (Table II). The solo group showed no differences in ND across trials (Table II).

*Mean Absolute Power (MAP)*

Between groups (Figure 3E), the dyad group demonstrated higher MAP (higher values indicate greater effort) compared to the solo group during the training trials *(Training 1: $\Delta$ = 1.27 [0.20, 2.35], $p$ = 0.024; Training 2: $\Delta$ = 1.53 [0.46, 2.60], $p$ = 0.009; Training 3: $\Delta$ = 1.35 [0.28, 2.43], $p$ = 0.017).* Likewise, the dyad group increased their MAP substantially during the training trials relative to baseline before returning to baseline performance at assessment and retention while the solo group showed no differences in MAP across trials (Figure 3F, Table II).

<u>*Sample Entropy (SampEn)*</u>

Between groups (Figure 3G), the dyad group had lower SampEn (lower values indicate less movement complexity) compared to the solo group at baseline (*Δ = -0.09 [-0.17, -0.01], p=0.029)*, but this difference was not apparent in subsequent trials. Within groups, the dyad group increased SampEn during the training trials relative to their baseline performance but only partially maintained this increase in the subsequent assessment and retention trials (Figure 3H, Table II). The solo group showed no change in SampEn (Figure 3H, Table II).

***Relationship between Motor/Cognitive Ability and Baseline Task Performance***

As shown in Figure 4, only RMSE showed significant correlations with clinical assessments: baseline RMSE showed strong ($r$ > 0.6) correlations with score on MoCA-EF ($r$ = -0.79), completion time on CT2 ($r$ = 0.78), score on FM-UE ($r$ = 0.77), and number of blocks on BBT ($r$ = 0.72). Given that greater RMSE indicates worse performance, these results revealed that better cognition, specifically better executive function, (high MoCA-EF score, low CT2 completion time) and worse motor function (low FM-UE score, fewer blocks in BBT) were associated with lower error on the task.

Although they did not reach significance, we noted a few moderate ($r$ > 0.4) correlations for SampEn (Figure 4). Namely, baseline SampEn showed moderate correlations with CT2 ($r$ = 0.47) and SWM ($r$ = 0.54). The direction of these relationships would suggest that greater movement complexity at baseline may be associated with better executive functions, particularly mental flexibility and selective attention, as well as intact light touch sensation.

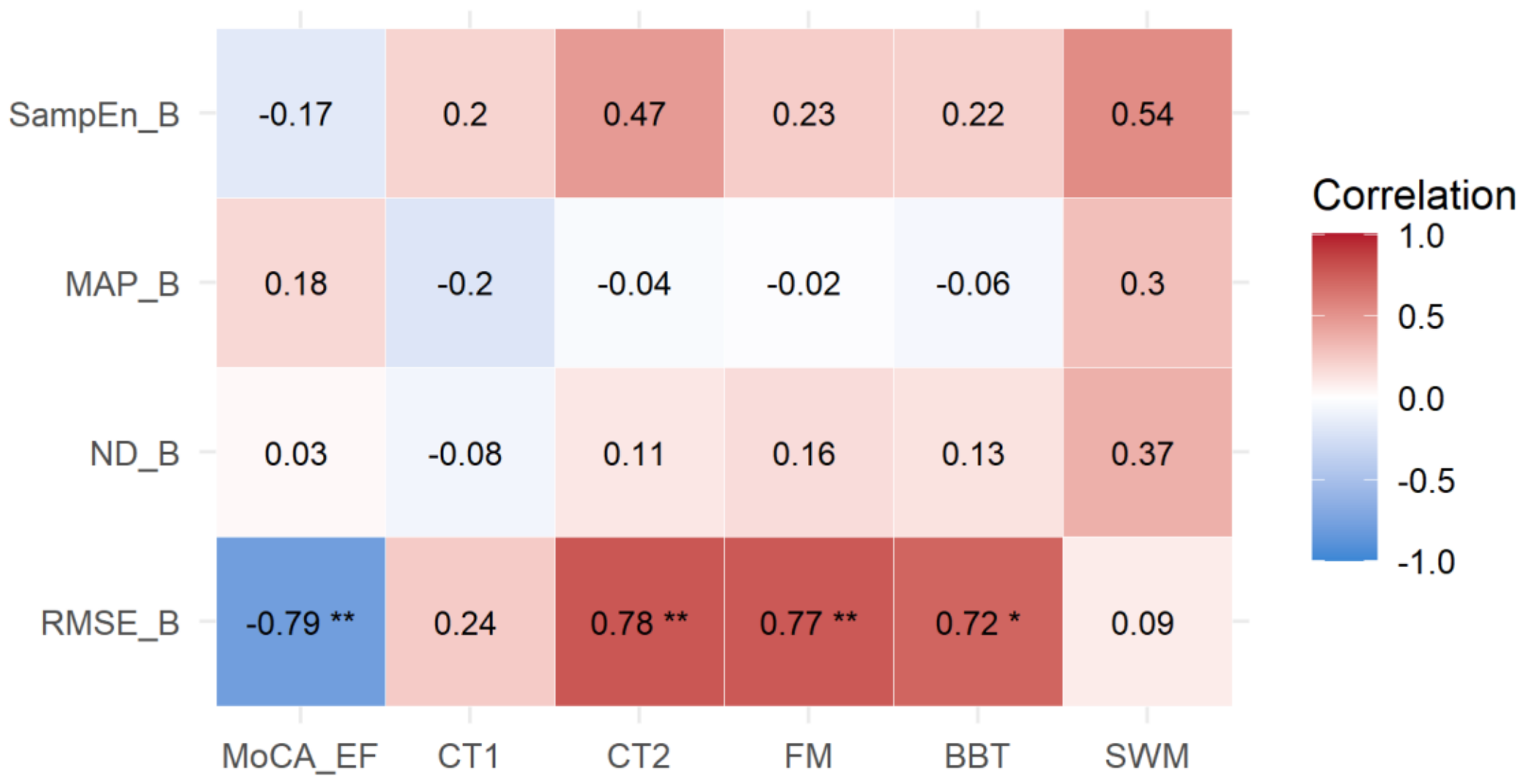

***Figure 4:*** *Heatmap of Pearson correlation coefficients between baseline task performance metrics and clinical assessments. Significant correlations are marked (p<0.05: *; p<0.01: **). Note that MoCA-EF, FM, BBT, and SWM are oriented such that a more positive value indicates better function; CT1 and CT2 are reported as completion time so a more positive value indicates worse function.*

## Discussion

We investigated how individuals with chronic stroke learned a motor task in solo and dyad conditions. We analyzed task performance during interaction, motor learning, and 1-week retention of learned task performance. While other studies have investigated motor learning in robot-based haptic dyads for healthy adults (Beckers *et al.*, 2020; Ganesh *et al.*, 2014; Ivanova *et al.*, 2022; Mireles *et al.*, 2017; Waters *et al.*, 2026), few studies have investigated dyadic training for neurological populations (Vianello *et al.*, 2025; Waters and Johnson, 2024), The studies that have investigated haptic dyads for post-stroke participants have not reported motor learning (Küçüktabak *et al.*, 2026) or only reported same-day learning without retention (Short *et al.*, 2025; Waters *et al.*, 2024). No studies have investigated retention of motor learning in robot-based haptic dyads for participants with post-stroke motor, cognitive, and sensorimotor impairments.

### *The Tracking Task had Higher Cognitive Demand than Motor Demand*

The tracking task, which involved mapping horizontal robot movement to vertical movement on the screen, had a high cognitive challenge. The role of attention in motor learning and tracking performance is well documented (Song, 2019; Azar *et al.*, 2026). Visuomotor mapping further involves executive functions such as spatial working memory and selective attention (Anguera *et al.*, 2010; Simon and Bock, 2017). Accordingly, better cognitive scores, i.e. MoCA-EF and CT2, were associated with lower baseline error (Figure 4). This relationship was not as strong for CT1, which could suggest the task favors executive function over attention. These results are aligned with recent findings showing that post-stroke cognitive impairment is associated with worse performance on robot-based preview tracking (Azar *et al.*, 2026; Bui *et al.*, 2022). Paradoxically, correlation analyses also revealed that better motor scores, i.e. FM-UE and BBT, were associated with higher tracking error at baseline (Figure 4). From the literature, lower post-stroke motor impairment is correlated with better final visuomotor adaptation but not initial adaptation or time to adapt (Cluff, 2022). At baseline, participants were in the early stages of adaptation. The positive relationship between motor scores and tracking error, therefore, may reflect that participants with lower motor impairment explore the workspace during baseline more than participants with higher motor impairment. This exploration could be indicative of higher motor variability, which is important for motor learning (Lyons *et al.*, 2025; Wu *et al.*, 2014). Notably, we did not employ a trunk restraint, so participants with more severe motor impairments were free to employ compensation strategies using the trunk and/or proximal limb, which has been observed on similar robot-based tasks with post-stroke participants (Meng *et al.*, 2025). We also did not account for lesion side or handedness, which is related to visuomotor adaptation (Cluff, 2022; Schaefer *et al.*, 2009). Together, these results support that the pursuit tracking task has higher cognitive demand than motor demand but may have still been too motorically challenging for the participants with moderate to severe motor impairment.

### *Dyadic Training Provides Temporary, but Unsustained, Performance Improvement*

As expected, the post-stroke participants in the dyad group showed an instantaneous improvement during the interaction trials. During the training trials, dyad participants significantly decreased RMSE relative to baseline (Figure 3B). Improved motor performance, particularly tracking error, while connected to a more skilled partner is well documented in the literature for healthy adults (Beckers *et al.*, 2020; Ganesh *et al.*, 2014; Ivanova *et al.*, 2022; Waters *et al.*, 2026), and has even been shown for connection to a therapist during ankle tracking in post-stroke participants (Short *et al.*, 2025). Dyad participants also moved more during interaction, exhibiting higher ND values during the training trials relative to baseline (Figure 3D). The solo group, conversely, did not change their ND across trials (Figure 3D). This increase in RMSE and ND for dyad but not solo participants during training mirrors previous findings with healthy older adults (Waters *et al.*, 2026).

Unlike healthy older adults, however, dyad participants did not maintain their improved performance once the connection to the healthy partner was removed. Despite showing lower RMSE and higher ND during training compared to baseline, the dyad group did not show improved performance at assessment or retention relative to baseline (Figure 3B, Figure 3D). The solo group, however, did show lower RMSE at assessment relative to baseline but this improvement was not maintained at retention (Figure 3B). The solo group also did not show any change in ND across trials (Figure 3D). We further note that one of the solo participants, participant 22, displayed substantially better performance than all other participants (Figure 3B), which may have skewed these results. In fact, removing this participant from the model reduced the observed difference between baseline and assessment RMSE for the solo group *($\Delta$ = -5.19 [-11.05, 0.66], $p$ = 0.106).* Furthermore, removing participant 22 from the analysis revealed significantly lower RMSE for the dyad group relative to solo group at all three training trials *(Training 1: $\Delta$ = -6.26 [-11.65, -0.87], $p$ = 0.025; Training 2: $\Delta$ = -6.66 [-12.06, -1.27], $p$ = 0.018; Training 3: $\Delta$ = -7.85 [-13.24, -2.46], $p$ = 0.007)*. We note that this superior performance was not reflective of less motor impairment since participant 22 had low-to-moderate motor impairment (FM-UE = 27) and was less than 1-year post-stroke when he completed the study. While all participants were in the chronic stage of stroke and were no longer in the traditionally accepted "critical window" for heightened neuroplasticity, recent evidence suggests that neuroplasticity declines over a gradient extending past 1-year post-stroke (Ballester *et al.*, 2019). Participants in the early chronic stage (<18 months post-stroke), therefore, should have a greater capacity for motor learning than participants in the late chronic stage. Participant 22 was not the only participant in the early chronic stage, but he was the only one to demonstrate intact executive function, i.e. CT2 and MoCA-EF, (Table I), which was related to baseline error (Figure 4). We further note that participant 22 had a pontine stroke, which may predominantly affect descending motor pathways, depending on the location and extent of the lesion, while the other early chronic participants had strokes in the area around the middle cerebral artery and in the basal ganglia, which potentially affect brain networks involved in executive function and motor learning (Agrawal *et al.*, 2025; Russo *et al.*, 2021). In this context, participant 22's remarkable performance and learning on the task may be explained by his relatively low chronicity and high function on task-relevant cognitive domains.

Despite participant 22 demonstrating clear learning, the overall results suggest that most post-stroke participants did not significantly improve on the task in the given time frame. This lack of measurable improvement in tracking error and distance traveled could indicate that the task was too difficult for the given timeframe, possibly due to the speed and visuomotor mapping. More evidence is needed to understand how lesion location, chronicity, and cognitive deficit relate to performance and learning in this motor task.

***Dyadic Interaction Increases Effort and Movement Complexity during Training***

During interaction, post-stroke participants not only decreased their tracking error and increased their distance traveled, but they also increased their effort and movement complexity. In fact, dyad participants demonstrated significantly higher training MAP relative to their own baseline performance and relative to the solo group (Figure 3E). MAP was consistently high during training with no clear changes across interaction trials. Sustained effort during interaction is well supported in the haptic dyad literature: more information about the partner via stiffer haptic connection and higher interaction forces has been shown to increase coordination in haptic dyad tasks, and counterintuitively, increasing coordination between partners over time does not lead to decreased interaction forces (Pena-Perez *et al.*, 2023; Takai *et al.*, 2023). These results demonstrate that not only do post-stroke participants not slack during the interaction trials, confirming previous findings (Waters *et al.*, 2024), but they exert greater effort than a solo condition.

The dyad participants also showed significantly higher SampEn during training compared with their baseline trials (Figure 3H), representing increased movement complexity. While this increase was not fully maintained, we note that the dyad group's SampEn at assessment and retention fell between their baseline and training values (Figure 3H). The solo group, conversely, did not change SampEn over time. Higher movement entropy is associated with better motor learning (Brink *et al.*, 2025), and individuals with chronic stroke demonstrate decreased entropy during reach-to-grasp tasks (Sethi *et al.*, 2013). Increasing movement entropy during training, therefore, could lead to improved motor learning given sufficient time to learn a task, but more evidence is needed. It is possible, however, that participants with significant sensory impairments would not be able to interpret the haptic signals and therefore not benefit from this type of intervention. While we were not able to test this hypothesis, we anecdotally observed that participant 40, who was the only participant to demonstrate measurable impairment in proprioception and had “deep pressure sensation only” in parts of his hand (Table I), had substantially lower MAP and SampEn during training than the other dyad participants, and the other dyad participant with severely impaired light touch sensation, participant 56, was the next lowest (Figure 3F, Figure 3H).

## Conclusion

We present a pilot study investigation of individuals with chronic stroke learning a complex tracking task in solo and dyad conditions. The task was too difficult to reveal clear learning or retention patterns, but evidence during training trials suggests that interacting with a partner can increase motor performance, effort, and movement complexity. Clinical characteristics such as lesion location, cognition, sensorimotor function, and chronicity may influence whether participants benefit from a haptic dyad intervention. Future work should include more participants from across the spectrum of post-stroke motor and cognitive impairments.

## Acknowledgement

We thank Joana Opong-Duah for help with study administration as well as Zalika Fletcher, Kelsey Shores, and the entire staff at the Penn Medicine Clinical Simulation Center for their assistance with research facilities.

For funding support, we acknowledge the Department of Physical Medicine & Rehabilitation, the Department of Mechanical Engineering and Applied Mechanics, and the General Robotics, Automation, Sensing and Perception (GRASP) Laboratory at the University of Pennsylvania. This material is based upon work supported by the National Science Foundation Graduate Research Fellowship Program under Grant No. DGE-1845298. Any opinions, findings, and conclusions or recommendations expressed in this material are those of the author(s) and do not necessarily reflect the views of the National Science Foundation. Research reported in this publication was also funded by the Eunice Kennedy Shriver National Institute Of Child Health & Human Development of the National Institutes of Health under Award Numbers F31HD116597 and R42HD104325. The content is solely the responsibility of the authors and does not necessarily represent the official views of the National Institutes of Health. Dr. Michelle Johnson and Dr. Rochelle Mendonca are co-founders of Recupero Robotics LLC, which may have a financial interest in this work.

A generative AI tool (Microsoft 365 Copilot, version no. 2.20260811.56.0 with enterprise data protection) was used to assist in writing analysis code for data manipulation and figures in Python and R. All output was thoroughly reviewed by the authors.